\documentclass[11pt]{article}
\usepackage[final]{acl}
\usepackage{xurl}%
\usepackage{times}
\usepackage{latexsym}
\usepackage[T1]{fontenc}
\usepackage[utf8]{inputenc}
\usepackage{microtype}
\usepackage{graphicx}
\usepackage{booktabs}

\usepackage[frozencache,cachedir=minted]{minted}
\setminted{style=trac,fontsize=\small}
\newminted{sparql}{escapeinside=``,vspace=4pt}
\newmintinline{sparql}{escapeinside=``}
\usepackage{multirow}
\usepackage{multicol}
\usepackage{enumitem}
\usepackage{threeparttable}
\usepackage{calc}
\usepackage{cleveref}
\usepackage{float}
\usepackage{algorithm}
\usepackage{algpseudocode}
\usepackage{xcolor}
\usepackage{subcaption}
\usepackage{afterpage}

\newcommand{\website}[1]{\href{https://grisp.cs.uni-freiburg.de}{\nolinkurl{grisp.cs.uni-freiburg.de}}}
\newcommand{\material}{\href{https://ad-publications.cs.uni-freiburg.de/grisp}{\nolinkurl{ad-publications.cs.uni-freiburg.de/grisp}}}
\newcommand{\git}{\href{https://github.com/ad-freiburg/grasp}{\nolinkurl{github.com/ad-freiburg/grasp}}}

\newcommand{\wdq}[1]{\href{https://www.wikidata.org/wiki/Q#1}{wd:Q#1}}

\newcommand{\wdtp}[1]{\href{https://www.wikidata.org/wiki/Property:P#1}{wdt:P#1}}

\begin{document}

\title{GRISP: Guided Recurrent IRI Selection over SPARQL Skeletons}

\author{Sebastian Walter \and Hannah Bast \\
  University of Freiburg, Department of Computer Science \\
  Freiburg im Breisgau, Germany \\
  \texttt{\{swalter,bast\}@cs.uni-freiburg.de}}

\maketitle

\begin{abstract}
We present \textbf{GRISP} (\textbf{G}uided \textbf{R}ecurrent \textbf{I}RI Selection over \textbf{SP}ARQL Skeletons), a novel SPARQL-based question-answering method over knowledge graphs using a fine-tuned small language model (SLM). Given a natural-language question, the method uses the SLM to generate natural-language SPARQL query skeletons, and then resolves their natural-language placeholders using a backtracking search that retrieves and re-ranks knowledge graph items under knowledge graph constraints. The SLM is jointly trained on skeleton generation and re-ranking data generated from question-query pairs. We evaluate the method on common Wikidata and Freebase benchmarks, and achieve better results than other strong fine-tuning-based methods in a comparable setting.
\end{abstract}

\section{Introduction}
\label{sec:intro}

Many large knowledge graphs (KGs) are available in RDF (the Resource Description Framework), including Freebase \cite{freebase}, Wikidata \cite{wikidata}, UniProt \cite{uniprot}, and OpenStreetMap \cite{osm}. This has driven growing interest in knowledge graph question-answering (KGQA) methods that generate queries in SPARQL, the standard query language for RDF data. At the same time, the recent emergence of agentic capabilities in large language models enabled the first zero-shot approaches for SPARQL-based KGQA, which need neither fine-tuning nor in-context examples. Systems like SPINACH \cite{spinach} and GRASP \cite{grasp} interactively explore and query the knowledge graphs in a human-like fashion and achieve state-of-the-art results on some of the KGQA benchmarks. However, they are typically slow and rely on large, expensive-to-run models. Fine-tuning the underlying model is also not feasible without significant hardware resources. If question-query pairs are available, methods based on fine-tuning small language models (SLMs) can outperform these agentic zero-shot systems in generation quality and speed, while needing less hardware. See \citet{grasp} for an overview. In this work, we present GRISP, a SPARQL-based KGQA method that follows a generate-then-retrieve approach. GRISP uses a fine-tuned SLM to generate natural-language SPARQL skeletons, in which natural-language labels are used instead of concrete knowledge graph identifiers (IRIs), and then resolves these placeholders using a backtracking search combined with a retrieve-and-re-rank procedure that uses the same SLM for re-ranking.

Let us look at GRISP answering the question \emph{``Where was the director of Pulp Fiction born?''} over Wikidata. GRISP processes it in two stages.

\paragraph{Stage 1: Skeleton Generation.}
GRISP uses beam search over the fine-tuned SLM to generate multiple SPARQL \emph{skeletons} for the question. A skeleton is a SPARQL query in which every IRI is replaced by a natural-language placeholder, written as \sparqlinline|PH:"..."|. GRISP discards duplicates and unparseable outputs, merges skeletons that have the same structure and differ only in their placeholder labels, and keeps the top candidates. For our example, the top-ranked skeleton might be

\begin{sparqlcode}
SELECT ?place WHERE {
  PH:"Pulp Fiction" PH:"director" ?d .
  ?d PH:"place of birth" ?place .
}
\end{sparqlcode}

\noindent Using multiple skeletons makes the method robust against variations in query structure (e.g.,~a competing skeleton might use the inverse relation \sparqlinline|?d PH:"directs" PH:"Pulp Fiction"|), and provides fallbacks if a skeleton cannot be fully~resolved.

\paragraph{Stage 2: Placeholder Resolution.}
For each skeleton, GRISP resolves the placeholders one after the other. At every placeholder, it performs a \emph{retrieve-and-re-rank} step: it queries the appropriate KG index (entity or property, depending on the syntactic position) with the placeholder label, collects the top alternatives together with a \emph{None} option, and has the SLM re-rank them. The top-ranked alternative replaces the~placeholder.\looseness=-1

In our example, \sparqlinline|PH:"Pulp Fiction"| is resolved against the entity index, yielding alternatives such as the 1994 film (\wdq{104123}) and the unrelated common-noun item \emph{pulp fiction} (\wdq{351718}). The SLM re-ranks them and selects the film. The partially resolved skeleton is now

\begin{sparqlcode}
SELECT ?place WHERE {
  wd:Q104123 PH:"director" ?d .
  ?d PH:"place of birth" ?place .
}
\end{sparqlcode}

\paragraph{KG Guidance.}
Before retrieving matches for the next placeholder, GRISP uses the already resolved part of the skeleton to constrain the candidate set: it rewrites the triple patterns connected to the placeholder into a single-column SPARQL query, executes it, and restricts the index search to the returned IRIs. For \sparqlinline|PH:"director"|, this means~\mbox{executing}

\begin{sparqlcode}
SELECT DISTINCT ?p WHERE {
  wd:Q104123 ?p ?d .
}
\end{sparqlcode}

\noindent so that only properties attached to \wdq{104123} are returned, e.g.,~\wdtp{57} (director), \wdtp{162} (producer), \wdtp{58} (screenwriter). The re-ranker picks \wdtp{57}. The same mechanism applies one step later: the constraining query for \sparqlinline|PH:"place of birth"| now chains through the resolved director,

\begin{sparqlcode}
SELECT DISTINCT ?p WHERE {
  wd:Q104123 wdt:P57 ?d .
  ?d ?p ?place .
}
\end{sparqlcode}

\noindent which restricts the retrieval to properties of that specific person. The re-ranker selects \wdtp{19}.

\paragraph{Backtracking.}
If re-ranking places \emph{None} on top, or the constrained retrieval returns no candidates, GRISP backtracks to the previous placeholder, drops the alternative that led there, and re-ranks the remaining alternatives of that placeholder. For instance, had the re-ranker initially selected \wdq{351718} instead of the film, the next constrained search for \sparqlinline|PH:"director"| would have returned no plausible properties, \emph{None} would have moved to the top, and GRISP would have backtracked to pick~the~film.

\paragraph{Validation and Final Answer.}
Once all placeholders are resolved, GRISP executes the SPARQL query and checks that it is \emph{valid}, i.e.,~that it parses, executes, and returns a non-empty result. Otherwise it backtracks one step. In our example, the query returns Knoxville (\wdq{185582}), the birthplace of Quentin Tarantino (\wdq{3772}). If a skeleton's resolution tree is exhausted, GRISP moves on to the next skeleton from Stage 1, and returns the first one that yields a valid result.

\subsection*{Core Contributions}
1. We present a novel SPARQL-based knowledge graph question-answering method for arbitrary knowledge graphs, called GRISP. The method is centered around fine-tuning an SLM to generate SPARQL skeletons and then resolve the placeholders to knowledge graph items. The main novelty lies in the placeholder resolution, which performs a backtracking search with a retrieve-and-re-rank operation under knowledge graph constraints for each placeholder.\\[1mm]
2. We evaluate the method on multiple benchmarks for the Freebase and Wikidata knowledge graphs, achieving better results than other strong fine-tuning-based methods in a comparable setting, while being much faster than agentic zero-shot~methods.\\[1mm]
3. We perform component studies and ablations to investigate how GRISP's mechanisms affect its performance. We also compare GRISP to GRASP, a recent state-of-the-art agentic KGQA method.\\[1mm]
4. We release our code as open source under a permissive license. It is part of the GRASP repository at \git{}, a demo is available at \website{}, and our evaluation including model checkpoints, benchmarks, and predictions is at \material{}.

\section{Related Work}
\label{sec:related}

KGQA methods broadly fall into two families. Methods in the first family traverse the knowledge graph and return the answers they find, e.g.,~Think-on-Graph \cite{tog}. They typically take the topic entities of the question as input. Recent methods of this family train an SLM agent with reinforcement learning (RL), e.g.,~Backjump-on-Graph \cite{bog}, GraphWalker \cite{graphwalker}, and the agent of \citet{ispkgr}, which can also issue SPARQL queries during traversal.

Methods in the second family perform semantic parsing and return a structured query, which is executed to obtain the answer. In this work we focus on this family. Two sub-categories are particularly relevant for GRISP: fine-tuning-based methods that train a (typically small) language model on question-query pairs, and agentic zero-shot methods in which a large language model explores the knowledge graph step by step to draft a query. Fine-tuning-based methods include DecAF \cite{decaf}, GETT-QA \cite{gettqa}, Pangu \cite{pangu}, WikiSP \cite{wikisp}, ChatKBQA \cite{chatkbqa}, RGR-KBQA \cite{rgrkbqa}, CompKBQA \cite{compkbqa}, and EVER \cite{ever}.

On some benchmarks, several of the methods above report F\textsubscript{1}-scores up to 9 percentage points above those we report for GRISP. However, these scores are not directly comparable to ours, because setting, metric, or base model differ. For example, the RL-trained agents are given the topic entities, and their F\textsubscript{1}-scores do not always require exact matches with the gold answers. We demonstrate this with GraphWalker, which defines its F\textsubscript{1}-score as the best token overlap between any predicted and any gold answer. When we rerun it on 200 random WebQuestionsSP questions, it reaches 89.1 under this definition, but only 42.2 under our metric, compared to 79.9 for GRISP (see~\Cref{app:graphwalker}).

For a fair comparison, we therefore have to run and evaluate a method ourselves, under the same conditions as GRISP. This requires its code, which is not available for some recent methods that report strong results, e.g.,~CompKBQA and EVER. As baselines, we choose ChatKBQA and WikiSP, which release their code and which we rerun as described in \Cref{sec:eval}. ChatKBQA achieves strong results on Freebase benchmarks and operates in the same generate-then-retrieve setting as GRISP. WikiSP is one of the few methods that focus on the larger and more popular Wikidata knowledge graph. We prefer WikiSP over GETT-QA, which also targets Wikidata, because WikiSP is more recent and reports strong results with a base model of similar size to ours on WikiWebQuestions and QALD-7, two of our main benchmarks. Most others, including ChatKBQA, still evaluate only on Freebase despite it having been read-only since 2015 and discontinued in 2016. We describe ChatKBQA and WikiSP in more detail in the following.\looseness=-1

ChatKBQA \cite{chatkbqa} generates S-expression skeletons and converts them to SPARQL for execution after retrieval, whereas GRISP directly generates SPARQL skeletons. After skeleton generation, ChatKBQA retrieves candidate entities via a lexical lookup using FACC1 entity mentions, and resolves properties directly from the skeleton placeholder labels. If direct resolution fails, it falls back to searching within the 2-hop neighborhood of retrieved entities and re-ranking candidates using SimCSE \cite{simcse}. The two methods differ mainly in the second stage. ChatKBQA does a one-shot lookup for entities and label-based resolution for properties, while GRISP iteratively resolves all placeholders via retrieve-and-re-rank. Resolving iteratively allows GRISP to apply knowledge graph constraints and to backtrack from wrong choices.

WikiSP \cite{wikisp} is a retrieve-then-generate method: it first retrieves entities using ReFinED \cite{refined} fine-tuned for Wikidata, and then generates a SPARQL query using the retrieved entities. For properties, WikiSP generates natural-language property labels in the SPARQL output and resolves them to Wikidata property IDs via an exact label match, falling back to the Wikidata search API if that fails.
Both ChatKBQA and WikiSP are closely tailored to their respective knowledge graphs and even to specific benchmarks, which limits generality. For example, ChatKBQA uses Freebase-specific FACC1 mentions for entity linking, a tuned S-expression parser per benchmark, and many hard-coded Freebase-specific SPARQL queries and optimizations in its code. WikiSP uses a Wikidata-tuned entity linker and the Wikidata search API. In contrast, we design \emph{and} implement GRISP to be knowledge-graph agnostic. Only the search indices need to be configured per knowledge graph, a one-time setup decoupled from the GRISP method itself.

On the agentic zero-shot side, the most relevant method for our work is GRASP \cite{grasp}, which we compare against in our evaluation. GRASP is knowledge-graph agnostic and uses an off-the-shelf large language model (smaller models perform significantly worse, see \citealp{grasp}) that interacts with the knowledge graph autonomously, interleaving reasoning steps with function calls to search, list, and execute SPARQL queries until it commits to a final query. Its main advantages over fine-tuning-based methods like GRISP are the much larger underlying model and the function-call loop, which allows it to adjust its query as many times as needed. The trade-offs are much higher latency and more hardware, since GRASP needs the larger model to perform well. We reuse GRASP's search indices for GRISP, see \Cref{sec:method}. For any knowledge graph, these search indices can be derived automatically by GRASP itself \cite{wikikgqa}.

\section{Method}
\label{sec:method}

\Cref{sec:intro} and \Cref{sec:related} sketch the overall working principle of GRISP and how it differs from related semantic-parsing and agentic methods. In the following, we describe the four mechanisms behind GRISP in more detail, namely skeleton generation with beam search, retrieve-and-re-rank, backtracking search, and KG guidance. Our two-stage generate-then-retrieve design is motivated by the observation that SLMs can perform semantic parsing reliably when IRIs are either known upfront or replaced by natural-language placeholders \citep{modernsempar, chatkbqa, rgrkbqa}. The resulting skeleton provides the structure for Stage 2, the placeholder resolution, which is GRISP's main novelty.

\paragraph{Skeleton Generation with Beam Search.} Beam search has been shown to improve KGQA performance, both for semantic parsing and knowledge graph traversal \cite{beamqa,tog,chatkbqa}. We also employ beam search with beam width $8$ and sampling during the skeleton generation phase to generate multiple SPARQL skeleton candidates using the SLM. We filter out duplicate skeletons and those that fail to parse. Skeletons that differ only in the wording of their placeholders are merged into one, which keeps up to three distinct wordings per placeholder in beam order. This way, differently worded copies of the same query do not take up several of the top-$3$ slots. We measure the effect in \Cref{tab:ablation}. From the remaining skeletons, we take the top-$3$ to the placeholder resolution stage.

\paragraph{Retrieve and Re-rank.} At each natural-language placeholder to be resolved, we perform an index search with the placeholder label, then re-rank the top-$10$ alternatives using the SLM and select the top-ranked alternative as replacement. Re-ranking is performed list-wise: all alternatives are presented to the SLM together and scored jointly in a single forward pass. Concretely, we assign a single-token identifier to each alternative and sort the alternatives by the next-token logit scores of their identifiers, similar to FIRST \cite{first}. Depending on the position of the placeholder in the query, we either search an entity index (containing subjects and objects) or a property index, using the search indices provided by GRASP. If a placeholder has several wordings from merged skeletons, we search them one after the other, and only use the next wording once the alternatives for the previous one are exhausted.
If the position inference fails, e.g.,~if a placeholder occurs in a \sparqlinline|FILTER| expression, we search both indices and re-rank the combined set of alternatives.
We always add a \emph{None} alternative to the search results before re-ranking in case no alternative matches. After re-ranking, all alternatives ranked above \emph{None} are kept on a stack for possible backtracking, while those ranked below \emph{None} are discarded. The top-ranked alternative is then popped from the stack and replaces the placeholder. When backtracking to a previous state, we re-rank the remaining alternatives in its stack rather than reusing the original order. Re-ranking costs only a single forward pass, and the original ordering below the top position may be unreliable since training supervises only that position (see the Training Data paragraph in \Cref{sec:eval}).

\paragraph{Backtracking Search.} The placeholder resolution stage can be viewed as a depth-first search over the tree of placeholder resolutions, with each node being a partially resolved skeleton and each edge one of the re-ranked alternatives for the next placeholder. To recover from a path in this tree that leads to a dead end or to an invalid SPARQL query, GRISP can go back to previous states. For example, this happens if GRISP re-ranks the \emph{None} alternative to the top~position.

\paragraph{KG Guidance.} To constrain the set of possible replacements for the current placeholder, we derive a single-column SPARQL query from the triple patterns connected to the current placeholder in the already resolved part of the SPARQL skeleton, execute it, and restrict the corresponding index search to the returned candidate IRIs. For example, this allows GRISP to respect constraints such as ``Only search within properties known to occur for female politicians born in Canada.'' What is already resolved around a placeholder depends on the order in which the placeholders are resolved. The order is configurable, and GRISP implements six orders, e.g.,~left to right or all entities before all properties (see \Cref{app:fillorder}).
This context-sensitive guidance reduces false positives during placeholder resolution and often reveals empty or invalid query paths before the skeleton is fully resolved. As a final instance of the same mechanism, once all placeholders are resolved, we execute the fully resolved SPARQL query against the target knowledge graph and backtrack if it fails to execute or returns an empty result. Note that in its current implementation, constrained search is skipped for placeholders occurring inside advanced SPARQL constructs like \sparqlinline|UNION|. In those cases we fall back to an unconstrained search. We do the same if the constraint query takes longer than one second or returns more than $2^{17}$ IRIs. This happens when little around the placeholder has been resolved so far, e.g.,~only a frequent property like \sparqlinline|wdt:P31|. The query then matches a large part of the knowledge graph and is too expensive to execute and process.

\section{Evaluation}
\label{sec:eval}

\begin{table}[t]
	\centering
	\caption{F\textsubscript{1}-score and average runtime per question in seconds of ChatKBQA and GRISP on Freebase and Wikidata benchmarks, both with Qwen2.5 7B. Baselines with their original base models are in \Cref{tab:baselines_original}.}
	\label{tab:main}
	\begin{threeparttable}
\setlength{\tabcolsep}{4pt}
\begin{tabular}{@{}l@{\hskip 0.4cm}cc@{\hskip 0.4cm}cc@{}}
\toprule
 & \multicolumn{2}{c@{\hskip 0.4cm}}{\textbf{ChatKBQA}} & \multicolumn{2}{c@{}}{\textbf{GRISP}} \\
\textbf{Benchmark} & F\textsubscript{1} & Time\tnote{1} & F\textsubscript{1} & Time \\
\midrule
\multicolumn{5}{@{}l}{Freebase} \\
\midrule
CWQ & 74.4 & 17 & 79.0 & 5.1 \\
WQSP & 76.1 & 15 & 79.5 & 4.0 \\
\midrule
\multicolumn{5}{@{}l}{Wikidata} \\
\midrule
QALD-7 & 34.7 & 53 & 45.3 & 4.5 \\
QALD-10 & 34.0 & 64 & 39.6 & 7.6 \\
WWQ & 73.0 & 4.1 & 80.3 & 3.5 \\
WWQ\textsubscript{dev} & 77.4 & 17 & 84.9 & 3.4 \\
\bottomrule
\end{tabular}
\begin{tablenotes}[flushleft]
\footnotesize
\item[1] Time measured on up to 100 random questions per benchmark, rerun on the same machine as GRISP (see \Cref{app:baselines})
\end{tablenotes}
\end{threeparttable}

\end{table}

\begin{table}[t]
	\centering
	\caption{F\textsubscript{1}-score and average runtime per question in seconds of WikiSP and GRISP on Wikidata benchmarks, both with Qwen2.5 7B.}
	\label{tab:additional}
	\begin{threeparttable}
\setlength{\tabcolsep}{4pt}
\begin{tabular}{@{}l@{\hskip 0.4cm}cc@{\hskip 0.4cm}cc@{}}
\toprule
 & \multicolumn{2}{c@{\hskip 0.4cm}}{\textbf{WikiSP}} & \multicolumn{2}{c@{}}{\textbf{GRISP}} \\
\textbf{Benchmark} & F\textsubscript{1} & Time & F\textsubscript{1} & Time \\
\midrule
\multicolumn{5}{@{}l}{Wikidata} \\
\midrule
QALD-7 & 41.2 & 0.8 & 45.3 & 4.5 \\
QALD-10 & 28.7 & 0.9 & 39.6 & 7.6 \\
WWQ & 72.6 & 0.7 & 80.3 & 3.5 \\
WWQ\textsubscript{dev} & 78.0 & 0.7 & 84.9 & 3.4 \\
\midrule
\multicolumn{5}{@{}l}{Additional Wikidata benchmarks} \\
\midrule
LC-QuAD 2.0 & 58.9 & 0.9 & 77.5 & 4.2 \\
SimpleQuestions & 80.1 & 0.6 & 91.1 & 2.7 \\
SPINACH\tnote{1} & 26.7 & 2.8 & 33.9 & 15 \\
QAWiki\tnote{1} & 42.5 & 1.3 & 47.9 & 9.1 \\
WDQL & 58.6 & 1.7 & 61.0 & 8.4 \\
\bottomrule
\end{tabular}
\begin{tablenotes}[flushleft]
\footnotesize
\item[1] Both systems trained on WDQL, as the benchmark has no train set
\end{tablenotes}
\end{threeparttable}

\end{table}

\paragraph{Benchmarks.} Our main benchmarks are WebQuestionsSP (WQSP) \cite{wqsp} and ComplexWebQuestions (CWQ) \cite{cwq} for Freebase, and QALD-7 \cite{qald7}, QALD-10 \cite{qald10}, and the test and dev sets of WikiWebQuestions (WWQ, WWQ\textsubscript{dev}) \cite{wikisp} for Wikidata. We report WWQ\textsubscript{dev} because \citet{wikisp} do so as well. It is also our validation set when training on WWQ and is used in our component and ablation studies. We also evaluate on five further Wikidata benchmarks: LC-QuAD 2.0 \cite{lcquad2}, SimpleQuestions \cite{simplequestions}, SPINACH \cite{spinach}, QAWiki \cite{qawiki}, and Wikidata Query Logs (WDQL) \cite{wdql}. See \Cref{app:benchmarks} for benchmark sizes and preprocessing.

\paragraph{Metric.} We evaluate all systems with the row-wise F\textsubscript{1}-score of \citet{grasp}, averaged over questions. We differ in one case: if the gold query is an ASK query, a prediction that is not an ASK query scores 0, whereas \citet{grasp} count any non-empty result as yes. The scores of GRASP on QALD in \Cref{tab:grisp_vs_grasp} therefore differ slightly from~theirs.

\begin{table*}[t]
	\centering
	\caption{F\textsubscript{1}-score and average runtime per question in seconds of GRASP (zero-shot) with Qwen2.5 72B and GPT-4.1 and of GRISP (fine-tuned) with Qwen2.5 7B on Freebase and Wikidata benchmarks. Bold and underline mark the best and second-best per column, i.e.,~the highest F\textsubscript{1}-score and the lowest runtime. Note that we only evaluate on samples where all methods provide predictions (GRASP evaluates on random subsets of at most 200 samples). For GRASP, we score its original predictions with our metric against the same knowledge graphs as GRISP, and take the runtimes from its original runs. Consequently, scores and runtimes for GRISP can differ from those in \Cref{tab:main,tab:additional}, and those for GRASP from the ones in \citet{grasp}.}
	\begin{threeparttable}
\setlength{\tabcolsep}{4pt}
\begin{tabular}{@{}l@{\hskip 0.3cm} c c@{\hskip 0.35cm} c c@{\hskip 0.35cm} c c@{\hskip 0.35cm} c c@{\hskip 0.35cm} c c@{\hskip 0.35cm} c c@{}}
\toprule
\textbf{System}
  & \multicolumn{2}{c@{\hskip 0.35cm}}{\textbf{CWQ}}
  & \multicolumn{2}{c@{\hskip 0.35cm}}{\textbf{WQSP}}
  & \multicolumn{2}{c@{\hskip 0.35cm}}{\textbf{QALD-7}}
  & \multicolumn{2}{c@{\hskip 0.35cm}}{\textbf{QALD-10}}
  & \multicolumn{2}{c@{\hskip 0.35cm}}{\textbf{WWQ}}
  & \multicolumn{2}{c@{}}{\textbf{SPINACH}}
  \\
  & F\textsubscript{1} & Time & F\textsubscript{1} & Time & F\textsubscript{1} & Time & F\textsubscript{1} & Time & F\textsubscript{1} & Time & F\textsubscript{1} & Time \\
\midrule
GRASP \scriptsize Qwen2.5 72B & 29.6\rlap{\tnote{2}} & 53 & 40.3 & 45 & \underline{70.5}\rlap{\tnote{2}} & 37 & \underline{60.0} & 47 & 68.4 & 46 & 28.9\rlap{\tnote{2}} & 132 \\
GRASP \scriptsize GPT-4.1 & \underline{47.0} & \underline{37} & \underline{52.1} & \underline{30} & \textbf{76.4} & \underline{17} & \textbf{71.9} & \underline{22} & \underline{75.4} & \underline{21} & \textbf{43.9} & \underline{45} \\
\midrule
GRISP \scriptsize Qwen2.5 7B & \textbf{82.1} & \textbf{5.4} & \textbf{79.7} & \textbf{4.2} & 45.3 & \textbf{4.5} & 38.9 & \textbf{8.6} & \textbf{80.4} & \textbf{3.6} & \underline{33.9}\rlap{\tnote{1}} & \textbf{15} \\
\bottomrule
\end{tabular}
\begin{tablenotes}[flushleft]
\footnotesize
\item[1] Trained on WDQL since SPINACH has no train set
\item[2] Not reported in \citet{grasp} but original predictions are available
\end{tablenotes}
\end{threeparttable}

	\label{tab:grisp_vs_grasp}
\end{table*}

\paragraph{Training Data.}

Given a training sample in the form of a question-query pair, we generate training data for skeleton generation by using its question and replacing all IRIs in the SPARQL query with a natural-language placeholder, which is randomly set to the IRI's main label or an alias. The definition of main label and alias is determined by the search index and varies between knowledge graphs, e.g.,~for Wikidata it is \sparqlinline|rdfs:label| and \sparqlinline|skos:altLabel|, respectively. For list-wise re-ranking we randomly choose an IRI from the query, then search in the index corresponding to the IRI's position, using the IRI's natural-language placeholder, to get the alternatives to re-rank for this IRI. Since the resolution order determines which placeholders are already resolved at inference, we fill in a random subset of the other IRIs in each training sample, so that a single model works with any resolution order. To make the re-ranker more robust, we perform four data augmentations for this stage. First, we randomly vary the search top-$k$, which determines the number of alternatives to re-rank. Second, we randomly drop the additional metadata shown for each alternative for disambiguation, which again is determined by the search index (e.g.,~\sparqlinline|schema:description|  for Wikidata). Third, we randomly drop the target alternative (the one matching the IRI) to make the model learn when to choose \emph{None}. Fourth, we randomly shuffle the alternatives to avoid position bias.
The SLM is jointly trained on both types of data using standard next-token cross-entropy loss for skeleton generation, and cross-entropy restricted to the identifier tokens corresponding to available alternatives for list-wise re-ranking.
The latter differs from FIRST, which uses a learning-to-rank loss for re-ranking, because we only have the target alternative available for supervision, and not a full ordering of all alternatives. Note that the data generation procedure is run for each sample and epoch. This way, the model sees different training data for the same sample in each epoch, which further mitigates overfitting. Data generation can happen either online (during training) or offline (precomputed for multiple epochs in advance). The offline option significantly speeds up training and helps with reproducibility.

\paragraph{Training Hyperparameters.} We use the instruction-tuned models from 0.5B to 7B parameters of the Qwen2.5 model family \cite{qwen2.5}. All results in this paper use the 7B variant unless stated otherwise. We train all models using LoRA \cite{lora} with $r=32$ and $\alpha=32$ on all linear layers for $4$ epochs ($1$ epoch on WDQL, which is much larger) with a base learning rate of $10^{-4}$, learning-rate warmup and cosine decay, a batch size of $8$, and early stopping. We train each model on a single NVIDIA H200 GPU (141 GB VRAM). Training times range from a few minutes on QALD-7 to about 14 hours on WDQL.

\paragraph{Inference Hyperparameters.} We use a temperature of $0.4$, a top-$p$ value of $0.9$, a beam width of $8$, and sampling for skeleton generation. We then keep the top-$3$ generated skeletons after merging for the placeholder resolution stage. For placeholder resolution, we set the search top-$k$ to $10$, and enable re-ranking, knowledge graph guidance, and backtracking as described in \Cref{sec:method}. We resolve placeholders triple by triple, with the entities of a triple before its property (see \Cref{app:fillorder} for other orders). We run GRISP and the baselines on the same server on a single NVIDIA RTX4090 GPU (24~GB VRAM).

\paragraph{Baselines.} We rerun ChatKBQA and WikiSP with the same base model as GRISP, against the same knowledge graph endpoints, and with the same metric. ChatKBQA is built for Freebase and WikiSP for Wikidata. On our main Wikidata benchmarks, we also evaluate a version of ChatKBQA ported to Wikidata. ChatKBQA's code contains a label fallback, which we do not use for the main results. Details on both baselines, the fallback, and more results are in \Cref{app:baselines}.

\paragraph{Results.} We compare GRISP to ChatKBQA in \Cref{tab:main} and to WikiSP in \Cref{tab:additional}. GRISP outperforms ChatKBQA and WikiSP on all benchmarks. On Freebase, it is 4.6 percentage points (p.p.) above ChatKBQA on CWQ and 3.4 p.p.\ on WQSP. On the main Wikidata benchmarks, neither baseline is consistently the better one, and GRISP is 4.1 to 7.3 p.p.\ above the better of the two. On the additional Wikidata benchmarks, its margin over WikiSP ranges from 2.4 p.p.\ on WDQL to 18.6 p.p.\ on LC-QuAD 2.0. The model trained only on WDQL also does well on benchmarks it has not seen in training. It reaches 33.9 on SPINACH, 47.9 on QAWiki, 63.4 on QALD-7, and 53.8 on QALD-10. On QALD, this is 18.1 and 14.2 p.p.\ above the models trained on the QALD training sets. Taking our stricter metric into account, these scores are on par with or better than those of \citet{wdql}, who use earlier versions of GRISP and WDQL.\looseness=-1

GRISP also outperforms both baselines with their original LLaMA2 base models (see \Cref{tab:baselines_original}). For these, we get 78.3 on CWQ and 77.9 on WQSP for ChatKBQA, even with its label fallback, and 73.0 on WWQ and 76.6 on WWQ\textsubscript{dev} for WikiSP.

GRISP's average runtime per question grows on harder benchmarks, where F\textsubscript{1} drops. On benchmarks like SPINACH, GRISP generates more involved skeletons, performs more retrieve-and-re-rank steps, and backtracks more often. Average runtimes range from 2.7s to 15s depending on benchmark complexity. WikiSP is faster with 0.6s to 2.8s per question, as it generates a single query with greedy decoding. ChatKBQA is slower on average with 4.1s to 64s per question: it resolves most questions quickly, but a few take minutes \mbox{(see~\Cref{app:baselines}).}

\begin{table*}[t]
	\centering
	\caption{F\textsubscript{1}-score and average runtime per question in seconds of GRISP on WWQ\textsubscript{dev} and QALD-10 for varying beam width \emph{n} (\Cref{tab:skeleton}), search top-\emph{k} (\Cref{tab:topk}), and model size (\Cref{tab:size}). All sub-tables use Qwen2.5 7B except for the size study. Bold and underline mark the best and second-best F\textsubscript{1}-score per benchmark within each sub-table.}
	\begin{minipage}[t]{0.32\textwidth}
		\centering
		\subcaption{Beam width \emph{n}.\label{tab:skeleton}}
		\setlength{\tabcolsep}{3pt}
\begin{tabular}{@{}l@{\hskip 0.25cm}cc@{\hskip 0.3cm}cc@{}}
\toprule
 & \multicolumn{2}{c@{\hskip 0.3cm}}{\textbf{WWQ\textsubscript{dev}}} & \multicolumn{2}{c@{}}{\textbf{QALD-10}} \\
\textbf{\emph{n}} & F\textsubscript{1} & Time & F\textsubscript{1} & Time \\
\midrule
16 & 84.4 & 4.1 & \textbf{40.2} & 9.7 \\
8 & \textbf{84.9} & 3.4 & \underline{39.6} & 7.6 \\
4 & \underline{84.8} & 2.9 & 38.8 & 6.3 \\
1 & 80.7 & 2.5 & 36.5 & 4.9 \\
\bottomrule
\end{tabular}

	\end{minipage}\hfill
	\begin{minipage}[t]{0.32\textwidth}
		\centering
		\subcaption{Search top-\emph{k}.\label{tab:topk}}
		\setlength{\tabcolsep}{3pt}
\begin{tabular}{@{}l@{\hskip 0.25cm}cc@{\hskip 0.3cm}cc@{}}
\toprule
 & \multicolumn{2}{c@{\hskip 0.3cm}}{\textbf{WWQ\textsubscript{dev}}} & \multicolumn{2}{c@{}}{\textbf{QALD-10}} \\
\textbf{top-\emph{k}} & F\textsubscript{1} & Time & F\textsubscript{1} & Time \\
\midrule
15 & \underline{84.7} & 3.6 & \underline{39.3} & 8.7 \\
10 & \textbf{84.9} & 3.4 & \textbf{39.6} & 7.6 \\
5 & 84.2 & 3.2 & 36.8 & 6.6 \\
1 & 82.4 & 3.0 & 32.2 & 6.7 \\
\bottomrule
\end{tabular}

	\end{minipage}\hfill
	\begin{minipage}[t]{0.32\textwidth}
		\centering
		\subcaption{Model size.\label{tab:size}}
		\setlength{\tabcolsep}{3pt}
\begin{tabular}{@{}l@{\hskip 0.25cm}cc@{\hskip 0.3cm}cc@{}}
\toprule
 & \multicolumn{2}{c@{\hskip 0.3cm}}{\textbf{WWQ\textsubscript{dev}}} & \multicolumn{2}{c@{}}{\textbf{QALD-10}} \\
\textbf{Size} & F\textsubscript{1} & Time & F\textsubscript{1} & Time \\
\midrule
7B & \textbf{84.9} & 3.4 & \textbf{39.6} & 7.6 \\
3B & \underline{83.4} & 3.3 & 36.2 & 5.7 \\
1.5B & 79.4 & 3.0 & \underline{38.6} & 5.7 \\
0.5B & 78.1 & 2.3 & 29.6 & 7.4 \\
\bottomrule
\end{tabular}

	\end{minipage}
\end{table*}

\paragraph{Comparison to GRASP.}

GRISP is much faster than agentic methods like GRASP (see \Cref{tab:grisp_vs_grasp}). It takes 3.6s to 15s per question, while GRASP takes 17s to 45s with GPT-4.1 and 37s to 132s with Qwen2.5 72B. On benchmarks whose query patterns can be learned from sufficient training data, GRISP clearly outperforms GRASP, which runs zero-shot. On CWQ, WQSP, and WWQ, GRASP with GPT-4.1 reaches 47.0, 52.1, and 75.4 compared to GRISP's 82.1, 79.7, and 80.4. When little training data is available, zero-shot agentic methods have the upper hand. On QALD-7 (100 training samples) and QALD-10 (412 training samples), GRASP with GPT-4.1 reaches 76.4 and 71.9 F\textsubscript{1} versus GRISP's 45.3 and 38.9. SPINACH is hard for all systems. GRISP, trained on the larger WDQL dataset, reaches 33.9 there, which is above GRASP with Qwen2.5 72B (28.9) but well below GRASP with GPT-4.1 (43.9). These results suggest that agentic methods benefit on such benchmarks from exploring the knowledge graph freely and revising their queries based on execution results, whereas GRISP cannot revise a skeleton once it is generated. GRISP remains a reasonable choice on them when answers have to come fast or when only a single consumer GPU is available.

\paragraph{Component and Ablation Studies.}

We run all studies on WWQ\textsubscript{dev}, which we use to choose our default hyperparameters. To see whether the findings carry over to a harder benchmark, we repeat the studies on the test set of QALD-10, which has no dev set.

We first study two important inference hyperparameters: the beam width for skeleton generation in \Cref{tab:skeleton} and the top-$k$ value for search in \Cref{tab:topk}. As expected, performance improves with increasing parameter values at first, but quickly saturates. On WWQ\textsubscript{dev}, F\textsubscript{1} is highest at a beam width of 8 and a top-$k$ of 10, which we therefore use by default. On QALD-10, going beyond these values changes F\textsubscript{1} by at most 0.6 p.p.\ in either direction, while runtime keeps growing. The top-$k$ value matters more on the harder QALD-10 benchmark, where a top-$k$ of 1 costs 7.4 p.p.\ compared to 2.5 p.p.\ on~WWQ\textsubscript{dev}.

Regarding the size of GRISP's underlying SLM, larger models generally perform better, as shown in \Cref{tab:size}. On WWQ\textsubscript{dev}, F\textsubscript{1} grows with every step in size, from 78.1 with 0.5B to 84.9 with 7B. On QALD-10, the 7B model is also the best, but the 1.5B model is ahead of the 3B model. The 0.5B model is already on par with WikiSP with Qwen2.5 7B on both benchmarks, and the 1.5B model outperforms WikiSP by 1.4 p.p.\ on WWQ\textsubscript{dev} and 9.9 p.p.\ on QALD-10.

Another inference setting is the order in which placeholders are resolved. Among the six orders, resolving triple by triple with entities before properties is the best on WWQ\textsubscript{dev}, so we use this order by default. It is also the best on WQSP and within 0.2 p.p.\ of the best on QALD-10. The order matters most on QALD-10, where the worst one is 3.8 p.p.\ below the default (see \Cref{app:fillorder}).

Next, we check the effect of removing core mechanisms from GRISP in \Cref{tab:ablation}. Removing any of them lowers F\textsubscript{1} on both benchmarks, but which one matters most depends on the benchmark. On WWQ\textsubscript{dev}, removing re-ranking costs most with 8.7 p.p., followed by beam search with 4.2 and knowledge graph guidance with 3.7. On QALD-10, knowledge graph guidance costs most with 6.3 p.p., followed by beam search with 3.1 and re-ranking with 2.6. Without knowledge graph guidance, the final query returns an empty result for 7.6\% of the questions on WWQ\textsubscript{dev} and 25.5\% on QALD-10. Skeleton merging, constrained search, and backtracking each cost at most 1.6 p.p. Here, \emph{w/o skeleton merging} keeps skeletons that differ only in the wording of their placeholders as separate candidates, \emph{w/o constrained search} disables using the SPARQL skeleton to constrain the index searches, and \emph{w/o KG guidance} disables both constrained search and the final SPARQL query validation.

We also report three minimal configurations of GRISP to see what beam search and re-ranking contribute on their own. \emph{All disabled} turns off beam search, re-ranking, constrained search, backtracking, and the final query validation. It generates a single skeleton and fills each placeholder with the top search result. \emph{Beam search only} and \emph{re-ranking only} each turn one of the mechanisms back on. Beam search alone gains just 0.8 p.p.\ on WWQ\textsubscript{dev} and 2.8 p.p.\ on QALD-10 over all disabled. Without the final query validation, GRISP has no way to reject a skeleton, so it always uses the first one. Re-ranking alone gains 7.8 p.p.\ on WWQ\textsubscript{dev} and 5.1 p.p.\ on QALD-10. In this configuration, re-ranking plays a similar role to the fine-tuned ReFinED entity linker in WikiSP, as both are trained to pick the entities for the query. With re-ranking only, GRISP is slightly above WikiSP on both benchmarks, while all disabled is clearly below (see \Cref{tab:additional}). Turning on all remaining mechanisms on top of re-ranking improves GRISP by another 6.5 p.p.\ on WWQ\textsubscript{dev} and 8.7 p.p.\ on QALD-10.\looseness=-1

Finally, we compare GRISP, which trains a single model on skeleton generation and re-ranking, against three alternative setups. In all three, the model that generates the skeletons is trained on skeleton generation only. They differ in the re-ranking: \emph{no re-ranker} skips it at inference, \emph{base re-ranker} uses the off-the-shelf Qwen2.5 7B model, and \emph{separate re-ranker} uses a second model trained only on re-ranking. Re-ranking with the off-the-shelf model is better than no re-ranking, by 2.2 p.p.\ on WWQ\textsubscript{dev} and 5.7 p.p.\ on QALD-10. Training the re-ranker adds another 5.2 p.p.\ on WWQ\textsubscript{dev}, but only 1.0 p.p.\ on QALD-10. Training two separate models brings no clear benefit over the single model, with 0.6 p.p.\ less on WWQ\textsubscript{dev} and 0.4 p.p.\ more on QALD-10. A single multi-task model is therefore preferable, as it saves the GPU memory needed for a second model.

\paragraph{GRISP's Behavior and Failure Modes.}
From interacting with GRISP ourselves, we find that it produces sensible SPARQL queries for simple questions (\href{https://qlever.dev/wikidata/AS04rd?exec=true}{``Where does Barack Obama live''}) almost always, for medium-difficulty questions (\href{https://qlever.dev/wikidata/ge61s4?exec=true}{``Albert Einstein's grandchildren, their parents, and their jobs''}) most of the time, and for hard questions (\href{https://qlever.dev/wikidata/GO1lSz?exec=true}{``All British monarchs''}) sometimes.
To see where GRISP's errors come from, we replace parts of it with oracles (see \Cref{tab:oracle}). With gold skeletons, GRISP reaches 98.2 on WWQ\textsubscript{dev} and 86.9 on QALD-10. Skeleton generation thus accounts for most of the lost F\textsubscript{1}: 13.3 of 15.1 p.p.\ on WWQ\textsubscript{dev} and 47.3 of 60.4 p.p.\ on QALD-10. GRISP's final query validation is a weak check for correctness. When GRISP returns a query, it has an F\textsubscript{1}-score of 0 in 10.8\% of the cases on WWQ\textsubscript{dev} and 46.0\% on QALD-10. However, a better validation would help little. The oracle validation, which accepts the first query with an F\textsubscript{1}-score of at least 0.9 among the top-$3$ skeletons, gains only 1.6 p.p.\ and 4.0 p.p., because a correct query is rarely among the candidates. GRISP's resolution mechanisms matter with gold skeletons as well, where disabling all of them costs 12.5 p.p.\ and 16.2 p.p. Backtracking is mostly a symptom of a wrong skeleton. On QALD-10, GRISP backtracks for 113 of the 384 questions, but only for 31 with gold skeletons. Backtracking also rarely rescues a question: of the 113 questions, 76 end without a query and only 7 are answered correctly. Accordingly, backtracking gains only 0.5 p.p., but it also costs only 0.5s per question (see \Cref{tab:ablation}). On WWQ\textsubscript{dev}, GRISP backtracks for only 21 of the 448 questions.

\section{Conclusion}

\begin{table}[t]
	\centering
	\caption{F\textsubscript{1}-score and average runtime per question in seconds of GRISP with Qwen2.5 7B on WWQ\textsubscript{dev} and QALD-10 for different mechanism ablations, minimal configurations, and training setups. Each mechanism ablation only removes that particular mechanism.}
	\setlength{\tabcolsep}{3pt}
\begin{tabular}{@{}l@{\hskip 0.25cm}cc@{\hskip 0.3cm}cc@{}}
\toprule
 & \multicolumn{2}{c@{\hskip 0.3cm}}{\textbf{WWQ\textsubscript{dev}}} & \multicolumn{2}{c@{}}{\textbf{QALD-10}} \\
\textbf{Variant} & F\textsubscript{1} & Time & F\textsubscript{1} & Time \\
\midrule
GRISP & 84.9 & 3.4 & 39.6 & 7.6 \\
\midrule
\multicolumn{5}{@{}l}{Mechanism ablations} \\
\midrule
w/o backtracking & 84.5 & 3.2 & 39.1 & 7.1 \\
w/o constrained search & 84.4 & 3.1 & 38.0 & 6.2 \\
w/o skeleton merging & 83.7 & 3.4 & 38.3 & 7.5 \\
w/o KG guidance & 81.2 & 2.8 & 33.3 & 4.9 \\
w/o beam search & 80.7 & 2.5 & 36.5 & 4.9 \\
w/o re-ranking & 76.2 & 4.9 & 37.0 & 10.4 \\
\midrule
\multicolumn{5}{@{}l}{Minimal configurations} \\
\midrule
re-ranking only & 78.4 & 2.2 & 30.9 & 3.2 \\
beam search only & 71.4 & 2.5 & 28.6 & 4.2 \\
all disabled & 70.6 & 1.8 & 25.8 & 2.7 \\
\midrule
\multicolumn{5}{@{}l}{Skeleton generation model with} \\
\midrule
separate re-ranker & 84.3 & 3.4 & 40.0 & 7.1 \\
base re-ranker & 79.1 & 4.0 & 39.0 & 6.9 \\
no re-ranker & 76.9 & 4.0 & 33.3 & 9.1 \\
\bottomrule
\end{tabular}

	\label{tab:ablation}
\end{table}

\begin{table}[t]
	\centering
	\caption{F\textsubscript{1}-score of GRISP and its \emph{all disabled} configuration from \Cref{tab:ablation} on WWQ\textsubscript{dev} and QALD-10 with generated skeletons, with gold skeletons, and with an oracle validation, which accepts the first query with an F\textsubscript{1}-score of at least 0.9 among those resolved from the top-$3$ skeletons.}
	\setlength{\tabcolsep}{4pt}
\begin{tabular}{@{}l@{\hskip 0.4cm}cc@{}}
\toprule
\textbf{Variant} & \textbf{WWQ\textsubscript{dev}} & \textbf{QALD-10} \\
\midrule
GRISP & 84.9 & 39.6 \\
w/ gold skeletons & 98.2 & 86.9 \\
w/ oracle validation & 86.5 & 43.6 \\
\midrule
all disabled & 70.6 & 25.8 \\
w/ gold skeletons & 85.7 & 70.7 \\
\bottomrule
\end{tabular}

	\label{tab:oracle}
\end{table}

We present GRISP, a novel KGQA method in a generate-then-retrieve setting based on fine-tuning an SLM on the tasks of SPARQL skeleton generation and re-ranking of knowledge graph items. The main novelty of GRISP is the placeholder resolution stage, which uses backtracking search to replace natural-language placeholders in SPARQL skeletons with IRIs, using a retrieve-and-re-rank operation under knowledge graph constraints for each placeholder. Our ablations confirm that this placeholder resolution stage is what sets GRISP apart from other methods. Within this stage, re-ranking and knowledge graph guidance contribute most, while backtracking, contrary to our expectations, adds little. We evaluate GRISP against other strong fine-tuning-based methods that operate in a comparable setting. It achieves consistent improvements over them on benchmarks for Freebase and Wikidata. GRISP is also much faster than current agentic methods and runs on a single consumer GPU. Given sufficient training data, it can match and even surpass their~performance.

GRISP stops as soon as it finds a valid query with a non-empty result, which is a weak check. For example, if the result for the question \emph{All British monarchs} does not contain Queen Elizabeth II, it is clearly wrong. However, our oracle validation shows that a better check alone would gain little, because for the questions GRISP gets wrong, usually none of its skeletons leads to a good answer. We therefore consider it an interesting direction for future work to combine a plausibility check for the result with a way to revise the skeletons in that case. Such a check requires knowledge about the world, and it is an open question how well an SLM can perform it.\looseness=-1

\section*{Limitations}

The main limitation of GRISP is that it relies on the availability of question-query training pairs, which limits applicability to knowledge graphs for which such data exists. In low-data regimes or on benchmarks with harder, more diverse queries, GRISP's performance falls behind state-of-the-art agentic methods, as we see on QALD-7, QALD-10, and SPINACH. Our oracle experiments suggest that skeleton generation becomes the dominant bottleneck in these regimes, while placeholder resolution stays closer to its ceiling. All reported results are from a single training run and a single inference run per configuration due to compute constraints. Runtimes also depend on many factors besides the method itself, such as server load, implementation, inference settings, the state of the SPARQL endpoint, and which queries a system happens to sample and execute. This makes them hard to measure in a fully comparable way, even on the same server. They show how the systems compare in order of magnitude, and small differences should not be over-interpreted.

As for potential risks, GRISP can in principle retrieve any information contained in the underlying knowledge graph, but its generated SPARQL queries and their execution results should not be blindly trusted. It is also naturally exposed to biases in the benchmark training data.

\section*{Acknowledgments} 

We did not use AI to do the core parts of the work. In particular, we did not use AI to implement the GRISP method or to write paper content from scratch. We did use AI coding assistants interactively to look over existing code and to write code for non-critical parts of this work, such as the demo webapp at \website{}. We also used AI to improve and polish our writing, and to rephrase existing paragraphs. We carefully checked AI-generated writing in all cases.

Funded by the Deutsche Forschungsgemeinschaft (DFG, German Research Foundation) -- Project-ID 499552394 -- SFB 1597.

We thank Luis Drayer for his insights and his help with rerunning WikiSP and ChatKBQA and with porting ChatKBQA to Wikidata.

\bibliography{grisp}

\appendix
\crefalias{section}{appendix}

\section{Benchmark Sizes and Preprocessing}
\label{app:benchmarks}

\paragraph{Preprocessing.} We host Freebase with Virtuoso \cite{virtuoso} and Wikidata with QLever \cite{qlever}. Some gold queries of the Wikidata benchmarks rely on behavior that is specific to the Wikidata Query Service, so we adapt them to run on QLever. For LC-QuAD 2.0, the Wikidata Query Service casts \texttt{YEAR(?x)} to a string inside \texttt{contains()}, while QLever does not. We rewrite the 348 test and 1,418 train gold queries with this pattern to \texttt{contains(str(YEAR(?x)), ...)}, and train and evaluate on the rewritten data. For SPINACH, we rewrite the gold queries from Wikidata Query Service syntax to standard SPARQL. For WDQL, we evaluate on a random subset of 1,000 test samples. We add the prefix declarations that the Wikidata Query Service predeclares to 213 of the gold queries. On all benchmarks, we score only the test samples whose gold query returns a result (see \Cref{tab:benchmarks}).\looseness=-1

\begin{table}[!t]
	\centering
	\caption{Train, dev, and test split sizes for all benchmarks used in this paper (\emph{split}), and the number of samples we use from each split (\emph{used}). The used train and dev samples are those that remain after conversion to GRISP training data. For benchmarks without a dev split, we hold out 10\% of the converted train samples as dev samples. The used test samples are those we score, i.e.,~those whose gold query returns a result. A dash indicates that the split does not exist or is not used. All benchmarks use English questions, but their SPARQL queries may ask for data in other languages, e.g.,~those of WDQL cover a wide range of languages.}
	\begin{threeparttable}
\newcommand{\benchsep}{\addlinespace[3pt]}
\setlength{\tabcolsep}{4pt}
\begin{tabular}{@{}l@{\hskip 0.25cm}l@{\hskip 0.25cm}rrr@{}}
\toprule
\textbf{Benchmark} & & \textbf{Train} & \textbf{Dev} & \textbf{Test} \\
\midrule
\multicolumn{5}{@{}l}{Freebase} \\
\midrule
\multirow{2}{*}{CWQ} & split & 27,639 & 3,519 & 3,531 \\
 & used & 27,605 & 3,517 & 3,273 \\
\benchsep
\multirow{2}{*}{WQSP} & split & 3,453 & - & 1,639 \\
 & used & 3,105 & 345 & 1,628 \\
\midrule
\multicolumn{5}{@{}l}{Wikidata} \\
\midrule
\multirow{2}{*}{QALD-7} & split & 100 & - & 50 \\
 & used & 90 & 10 & 46 \\
\benchsep
\multirow{2}{*}{WWQ\tnote{1}} & split & 2,431 & 454 & 1,431 \\
 & used & 2,428 & 454 & 1,407 \\
\benchsep
\multirow{2}{*}{SPINACH} & split & - & 155 & 165 \\
 & used & - & - & 148 \\
\benchsep
\multirow{2}{*}{WDQL} & split & 181,099 & 22,638 & 22,638 \\
 & used & 171,485 & 2,847 & 993 \\
\benchsep
\multirow{2}{*}{QALD-10} & split & 412 & - & 394 \\
 & used & 369 & 41 & 384 \\
\benchsep
\multirow{2}{*}{LC-QuAD 2.0} & split & 24,124 & - & 6,028 \\
 & used & 21,362 & 2,373 & 4,932 \\
\benchsep
\multirow{2}{*}{SimpleQ\tnote{2}} & split & 19,481 & 2,821 & 5,622 \\
 & used & 19,447 & 2,817 & 5,573 \\
\benchsep
\multirow{2}{*}{QAWiki} & split & - & - & 518 \\
 & used & - & - & 498 \\
\bottomrule
\end{tabular}
\begin{tablenotes}[flushleft]
\footnotesize
\item[1] WWQ\textsubscript{dev} in the main results refers to the dev split, of which 448 samples are scored
\item[2] SimpleQuestions
\end{tablenotes}
\end{threeparttable}

	\label{tab:benchmarks}
\end{table}

\section{Baseline Details}
\label{app:baselines}

\begin{table}[t]
	\centering
	\caption{F\textsubscript{1}-score of ChatKBQA and WikiSP with their original base models and of GRISP with Qwen2.5 7B on Freebase and Wikidata benchmarks. The versions with Qwen2.5 7B are in \Cref{tab:main,tab:additional}.}
	\begin{tabular}{@{}lccc@{}}
\toprule
\textbf{Benchmark} & \textbf{ChatKBQA} & \textbf{WikiSP} & \textbf{GRISP} \\
\midrule
CWQ & 78.3 & -- & 79.0 \\
WQSP & 77.9 & -- & 79.5 \\
WWQ & -- & 73.0 & 80.3 \\
WWQ\textsubscript{dev} & -- & 76.6 & 84.9 \\
\bottomrule
\end{tabular}

	\label{tab:baselines_original}
\end{table}

\paragraph{Baseline Setup.} All baselines in \Cref{tab:main,tab:additional} use Qwen2.5 7B and are scored with our metric against the same knowledge graphs as GRISP. We retrain ChatKBQA with this base model and run it without its label fallback (see below). We train WikiSP as in \citet{wikisp} on WWQ upsampled 5 times plus the same number of samples from Alpaca, for QALD-7 and QALD-10 additionally on the respective train set upsampled 20 times, and for SPINACH, QAWiki, and WDQL on the WDQL train set used for GRISP plus the same Alpaca samples. For LC-QuAD 2.0 and SimpleQuestions, we train it on the respective train set plus the same Alpaca samples. We base both reruns on existing projects. For ChatKBQA, we use the code at \url{https://github.com/Ludraaa/Bachelorarbeit}, which includes the version ported to Wikidata, and only add the label fallback for Wikidata described below. For WikiSP, our rerun follows the reproduction at \url{https://github.com/Ludraaa/Bachelorprojekt}.

\paragraph{Label Fallback.}\label{app:label_fallback}
ChatKBQA's code contains a label fallback that \citet{chatkbqa} do not describe. When a resolved query returns no result, its evaluation scripts replace every linked entity by a variable that matches any entity with the same English label, and execute the query again. The fallback is written for Freebase. For Wikidata, we adapt it as follows: each variable is bound via \texttt{rdfs:label} with language tag \texttt{en} at the start of the outermost WHERE clause, and queries with a VALUES clause or a subquery are skipped, because a variable cannot appear in a VALUES clause and the bindings would not reach into a subquery. In \Cref{tab:main} we run ChatKBQA without it. We consider this the fairer comparison, because the fallback is not part of the method as published, and because it is a generic retry by label that any system, including GRISP, could add. \Cref{tab:label_fallback} shows results with the fallback, on Wikidata with our adaptation of it. On Freebase, the fallback raises F\textsubscript{1} on CWQ from 74.4 to 77.1 and hardly changes it on WQSP, from 76.1 to 76.2. On Wikidata, F\textsubscript{1} changes by at most 0.6 p.p.: it stays at 34.7 on QALD-7, and goes from 34.0 to 34.4 on QALD-10, from 73.0 to 72.4 on WWQ, and from 77.4 to 77.7 on WWQ\textsubscript{dev}. The fallback thus helps on one Freebase benchmark only and does not transfer to Wikidata, where labels are far from unique. For example, more than 50 entities share the English label \emph{Jamaica}. GRISP stays ahead of ChatKBQA with the fallback on all benchmarks. The average runtime per question is about the same with and without the fallback on CWQ, WQSP, WWQ, and WWQ\textsubscript{dev}, and higher with it on QALD-7 and QALD-10, where a few slow questions account for the difference.\looseness=-1

\begin{table}[t]
	\centering
	\caption{F\textsubscript{1}-score on the full test sets (dev set for WWQ\textsubscript{dev}) and average runtime per question in seconds of ChatKBQA with Qwen2.5 7B without (w/o) and with (w/) its label fallback.}
	\begin{threeparttable}
\setlength{\tabcolsep}{4pt}
\begin{tabular}{@{}l@{\hskip 0.4cm}cc@{\hskip 0.4cm}cc@{}}
\toprule
 & \multicolumn{2}{c@{\hskip 0.4cm}}{\textbf{F\textsubscript{1}}\tnote{1}} & \multicolumn{2}{c@{}}{\textbf{Time}\tnote{2}} \\
\textbf{Benchmark} & w/o & w/ & w/o & w/ \\
\midrule
CWQ & 74.4 & 77.1 & 17.0 & 17.1 \\
WQSP & 76.1 & 76.2 & 15.1 & 15.1 \\
\midrule
QALD-7 & 34.7 & 34.7 & 52.6 & 56.7 \\
QALD-10 & 34.0 & 34.4 & 64.0 & 72.6 \\
WWQ & 73.0 & 72.4 & 4.1 & 4.2 \\
WWQ\textsubscript{dev} & 77.4 & 77.7 & 17.2 & 16.5 \\
\bottomrule
\end{tabular}
\begin{tablenotes}[flushleft]
\footnotesize
\item[1] On the full test sets. With the fallback, we resolve the same generations again, but stop after 120 seconds per question and then keep the query found without the fallback.
\item[2] Mean time in seconds per question for generation and resolution, on a random sample of 100 questions per benchmark (all 50 for QALD-7). Here, we resolve each question without and with the fallback from the same generations, without the 120-second limit.
\end{tablenotes}
\end{threeparttable}

	\label{tab:label_fallback}
\end{table}

\paragraph{Original Base Models.} \Cref{tab:baselines_original} shows both baselines with their original LLaMA2 base models. For ChatKBQA, we take the predictions released by its authors (LLaMA2 7B on WQSP, LLaMA2 13B on CWQ), resolve them again with the label fallback against the same knowledge graph endpoint as GRISP, and score them with our metric. For WikiSP, we rerun it with LLaMA2 7B, trained on WWQ as described above, and execute its queries against the same endpoint as GRISP. GRISP is ahead of both baselines on all four benchmarks.

\paragraph{Runtimes.} We measure all runtimes on the same server and against the same knowledge graph endpoints. GRISP and WikiSP are timed on the full test sets. ChatKBQA is rerun on this server for timing only, on a random sample of up to 100 questions per benchmark. WikiSP generates a single query with greedy decoding, while GRISP and ChatKBQA use beam search with 8 beams. For ChatKBQA, mean and median resolution time are far apart, with and without the label fallback, as it resolves most questions in about a second or less, but a few take minutes.\looseness=-1

\section{Placeholder Resolution Order}
\label{app:fillorder}

\begin{table*}[t]
	\centering
	\caption{F\textsubscript{1}-score of GRISP with Qwen2.5 7B on WWQ\textsubscript{dev}, QALD-10, and WQSP for six placeholder resolution orders. The first row is the default. Bold and underline mark the best and second-best per column.}
	\begin{tabular}{@{}l@{\hskip 0.4cm}ccc@{}}
\toprule
\textbf{Resolution order} & \textbf{WWQ\textsubscript{dev}} & \textbf{QALD-10} & \textbf{WQSP} \\
\midrule
per triple, entities before properties & \textbf{84.9} & \underline{39.6} & \textbf{79.5} \\
entities before properties & \underline{84.5} & \textbf{39.8} & 79.1 \\
left to right & 84.3 & 38.1 & 79.2 \\
right to left & 84.0 & 36.1 & 79.2 \\
properties before entities & 84.1 & 35.8 & \underline{79.4} \\
random & 84.3 & \textbf{39.8} & 79.3 \\
\bottomrule
\end{tabular}

	\label{tab:fillorder}
\end{table*}

We compare six orders in which GRISP resolves the placeholders of a skeleton (see \Cref{tab:fillorder}). \emph{Left to right} and \emph{right to left} follow the order of the placeholders in the skeleton, forwards or backwards. \emph{Entities before properties} resolves all entity placeholders first and then all property placeholders, and \emph{properties before entities} does the reverse. \emph{Per triple, entities before properties} goes through the skeleton triple by triple and resolves the entities of a triple before its property. When a property is resolved, the entities of its own triple are thus already known, in addition to the placeholders of earlier triples, and all of them constrain its candidates via the connected triple patterns. \emph{Random} resolves the placeholders in a random order, which is fixed per skeleton. All six orders use the same model, trained as described in \Cref{sec:eval}.

Per triple, entities before properties is the best order on WWQ\textsubscript{dev}, so we use it by default. It is also the best on WQSP. On these two benchmarks the order matters little, as the worst one is only 0.9 p.p.\ below the default on WWQ\textsubscript{dev} and 0.4 p.p.\ on WQSP. On QALD-10, entities before properties and random are 0.2 p.p.\ above the default, while right to left and properties before entities are 3.5 and 3.8 p.p.\ below.

\section{Comparison to GraphWalker}
\label{app:graphwalker}

We use the released GraphWalker-7B checkpoint (Qwen2.5 7B) and give it the gold topic entity. Since GraphWalker outputs answer labels rather than SPARQL, we compute its F\textsubscript{1} over labels, against the names of the entities that our gold SPARQL query returns. GraphWalker uses a second model to filter candidate relations, which is GPT-4o-mini in \citet{graphwalker}. We run it on 200 random questions per benchmark, once with GPT-4o-mini and once with GraphWalker-7B itself as this filter (see \Cref{tab:graphwalker}). GraphWalker-7B rarely returns a valid selection. Without one, GraphWalker continues with up to 20 of the candidate relations chosen at random. With either filter, GraphWalker is far below GRISP under our F\textsubscript{1}, although it is given the gold topic entity.

GraphWalker's own F\textsubscript{1} is the best token overlap between any predicted and any gold answer, so additional wrong answers are not penalized. Its exact match (EM) is lenient in a similar way: a question counts as correct if any predicted answer is equal to a gold answer after normalization, or if one of the two contains the other. Under its own F\textsubscript{1}, GraphWalker with GPT-4o-mini reaches 89.1 on WQSP and 85.0 on CWQ on the same questions, compared to 88.6 and 74.2 on the full test sets in \citet{graphwalker}.

\begin{table}[t]
	\centering
	\caption{F\textsubscript{1}-score of GraphWalker and GRISP on 200 random questions per Freebase benchmark, both with Qwen2.5 7B. GraphWalker is run with GraphWalker-7B (7B) and with GPT-4o-mini as relation filter. F\textsubscript{1} is our set-based score with exact matching over answer labels.}
	\begin{threeparttable}
\setlength{\tabcolsep}{2.5pt}
\begin{tabular}{@{}l@{\hskip 0.25cm}cc@{\hskip 0.3cm}cc@{\hskip 0.3cm}c@{}}
\toprule
 & \multicolumn{4}{c@{\hskip 0.3cm}}{\textbf{GraphWalker}\tnote{1}} & \textbf{GRISP} \\
Filter & \multicolumn{2}{c@{\hskip 0.3cm}}{7B} & \multicolumn{2}{c@{\hskip 0.3cm}}{GPT-4o-mini} & \\
\textbf{Benchmark} & F\textsubscript{1} & Own\tnote{2} & F\textsubscript{1} & Own\tnote{2} & F\textsubscript{1} \\
\midrule
CWQ & 28.7 & 74.6 & 35.6 & 85.0 & 76.7 \\
WQSP & 39.9 & 83.9 & 42.2 & 89.1 & 79.9 \\
\bottomrule
\end{tabular}
\begin{tablenotes}[flushleft]
\footnotesize
\item[1] Given the gold topic entity
\item[2] GraphWalker's own F\textsubscript{1}, the best token overlap between any predicted and any gold answer
\end{tablenotes}
\end{threeparttable}

	\label{tab:graphwalker}
\end{table}

On the same 200 random questions per benchmark, GraphWalker returns a median of 5 answers with either filter, while the gold answer set has a median size of 1. A set-based F\textsubscript{1} like ours penalizes these additional answers. If an oracle removed all wrong answers, GraphWalker would reach 68.6 on WQSP and 62.7 on CWQ with GraphWalker-7B as filter, and 75.9 and 77.8 with GPT-4o-mini.

\end{document}